\documentclass{PNAStwo}
\newcounter{defcounter}
\usepackage{graphicx}
\usepackage{caption}
\usepackage{amssymb,amsfonts,amsmath}
\usepackage{bm}
\usepackage{upgreek}
\usepackage{color}

\DeclareMathOperator*{\argmax}{arg\,max}
\DeclareMathOperator*{\arglocalmax}{arglocal\,max}
\DeclareMathSymbol{\upomega}{\mathord}{ugrf@m}{"21}
\DeclareMathOperator{\sinc}{sinc}

\newenvironment{myequation}{%
\addtocounter{equation}{-1}
\refstepcounter{defcounter}
\renewcommand\theequation{SI\thedefcounter}
\begin{equation}}
{\end{equation}}

\newenvironment{mysubequations}{%
\addtocounter{equation}{-1}
\refstepcounter{defcounter}
\renewcommand\theequation{SI\thedefcounter}
\begin{subequations}}
{\end{subequations}}

\newcommand{\figuremacroW}[4]{
	\begin{figure}[tbp]
		\centerline{\includegraphics[width=#4\textwidth]{#1}}
		\caption{\textbf{#2} - #3 \label{#1}}
	\end{figure}
}

\url{}
\copyrightyear{}
\issuedate{}
\volume{}
\issuenumber{}
\begin{document}
\clearpage

\title{Nonparametric Estimation of Band-limited Probability Density Functions}

\author{Rahul Agarwal\affil{1}{Johns Hopkins University, Baltimore, MD}, Zhe Chen \affil{2}{New York University School of Medicine, New York, NY 10016} \and Sridevi V. Sarma \affil{1}{}}

\contributor{}

\maketitle

\begin{article}

\begin{abstract}
In this paper, a nonparametric maximum likelihood (ML) estimator for band-limited (BL) probability density functions (pdfs) is proposed. The BLML estimator is consistent and computationally efficient. To compute the BLML estimator, three approximate algorithms are presented: a binary quadratic programming (BQP) algorithm for medium scale problems, a Trivial algorithm for large-scale problems that yields a consistent estimate if the underlying pdf is strictly positive and BL, and a fast implementation of the Trivial algorithm that exploits the band-limited assumption and the Nyquist sampling theorem (``\texttt{BLMLQuick}"). All three BLML estimators outperform kernel density estimation (KDE) algorithms (adaptive and higher order KDEs) with respect to the mean integrated squared error for data generated from both BL and infinite-band pdfs. Further, the \texttt{BLMLQuick} estimate is remarkably faster than the KD algorithms. 
Finally, the BLML method is applied to estimate the conditional intensity function of a neuronal spike train (point process) recorded from a rat's entorhinal cortex grid cell, for which it outperforms state-of-the-art estimators used in neuroscience.

\end{abstract}

\keywords{probability  density |  maximum likelihood | band limited | nonparametric | estimation | BLML | \texttt{BLMLQuick} | \texttt{BLML-BQP} | \texttt{BLMLTrivial}}

\abbreviations{BLML: Band limited maximum likelihood; pdf: probability density function;
KDE: Kernel density estimation; BQP: binary quadratic programming; MISE: mean integrated squared error; MNLL; mean normalized log-likelihood; KDE2nd, KDE6th, KDEsinc: 2nd and 6th order Gaussian and sinc Kernel density estimators; \texttt{BLMLQuick}, \texttt{BLML-BQP}, \texttt{BLMLTrivial}: BLML quick, trivial and BQP algorithms}

\section{Significance Statement}
Nonparametric estimation of probability densities has became increasingly important to make predictions about processes where parametrization is difficult. However, unlike parametric approaches, nonparametric approaches are often not optimal in the sense of maximizing likelihood, which would ensure several optimality properties of the estimator. In this study, a novel nonparametric density estimation technique that maximizes likelihood and that converges to the true pdf is presented. Simulations show that this technique outperforms and outruns sophisticated kernel density estimation techniques, and yields the fastest convergence rate to date.\\


\dropcap{W}hen making inferences from experimental data, it is often required to model random phenomena and estimate a pdf. A common approach is to assume that the pdf belongs to a class of parametric functions (e.g., Gaussian, Poisson), and then estimate the parameters by maximizing the data likelihood function. Parametric models have several advantages. First, they are often efficiently computable. Second, the parameters may be related back to physiological and environmental variables. Finally, ML estimates have nice asymptotic properties when the actual distribution lies in the assumed parametric class. However, if the true pdf does not lie in the assumed class of functions, large errors may occur, potentially resulting in misleading inferences.

When little is known a priori about the pdf, nonparametric estimation is an option. However, maximizing the likelihood function yields spurious solutions as the dimensionality of the problem typically grows with the number of data samples, $n$ \cite{Montricher1975}.  To deal with this, several nonparametric approaches penalize the likelihood function by adding a smoothness constraint. Such penalty functions have nice properties of having unique maxima that can be computed. However, when smoothness conditions are applied, the asymptotic properties of ML estimates are typically lost \cite{Montricher1975}.
 
Other methods for nonparametric estimation assume that the pdf is a linear combination of scaled and stretched versions of a single kernel function  \cite{Rosenblatt1956,Parzen1962,Peristera2008,Scaillet2004}. These methods fall under kernel density (KD) estimation, which have been studied for decades. However, choosing an appropriate kernel is still a tricky and often an arbitrary process \cite{Park1990}. Additionally, even the best KD estimators \cite{Park1990,Park1992,Hall1991,Sheather1996} have slower convergence rates ($ \mathcal{O}_p(n^{-4/5})$ , $\mathcal{O}_p(n^{-12/13})$ for the second and sixth-order Gaussian kernels, respectively) than parametric ML estimation ($\mathcal{O}_p(n^{-1})$) with respect to the mean integrated squared error (MISE)\cite{Kanazawa1993}.

Finally, some approaches require searching over nonparametric sets for which a maximum likelihood estimate exists. Some cases are discussed in \cite{Carandoa2009,Coleman2010}, wherein the authors construct maximum likelihood estimators for unknown but Lipschitz continuous pdfs. Although Lipschitz functions display desirable continuity properties, they can be non-differentiable. Therefore, such estimates can be non-smooth, but perhaps more importantly, they are not efficiently computable as a closed-form solution cannot be derived \cite{Carandoa2009,Coleman2010}. 

This paper presents a case where a nonparametric ML estimator {\it exists}, is efficiently computable, consistent and results in a smooth pdf. The pdf is assumed to be band-limited (BL), which has a finite-support in the Fourier domain. The BL assumption essentially can be thought of as a smoothness constraint. However, the proposed method does not require penalizing the likelihood function to guarantee the existence of a global maximum, and therefore may preserve the asymptotic properties of ML estimators (i.e. consistency, asymptotic normality and efficiency). The BLML method is first applied to surrogate data generated from both BL and infinite-band pdfs, where in both cases it outperformes all tested KD estimators (including higher order kernels) both in convergence rate and computational time. Then the BLML estimator is applied to the neuronal data recorded from a rat's entorhinal cortex ``grid cell" and is shown to outperform state-of-the-art estimators used in neuroscience.


\section{The BLML Estimator}
We begin with a description of the BLML estimator in the following theorem.

\begin{theorem}\label{BLML_thm} 
Consider $n$ independent samples of an unknown BL pdf, $f(x)$, with assumed cut-off frequency $f_c.$ Then the BLML estimator of $f(x)$ is given as:
\begin{equation}\label{BLML_eq}
\hat{f}(x)=\left(\frac{1}{n}\sum_{i=1}^n\hat{c}_i\frac{\sin(\pi f_c(x-x_i))}{\pi(x-x_i)}\right)^2,
\end{equation}
where $\bold{\hat{c}}\triangleq \left[ \hat{c}_1,  \cdots , \hat{c}_n \right]^T$ and
\begin{equation}\label{BLML_eq2}
\hat{\bold{c}}=\argmax_{ \bm{\rho_n}(\bold{c)=0} }\left(\prod_{i=1}^{n} \frac{1}{c_i^2}\right).
\end{equation}
Here $\rho_{ni}(\bold{c})\triangleq \frac{1}{n}\sum_{j=1}^n c_js_{ij}-\frac{1}{c_i}$ $\forall~i=1,\cdots,n$ and \\
$s_{ij}\triangleq \frac{\sin(\pi f_c(x_i-x_j))}{\pi(x_i-x_j)} \ \ \  \forall~i,j=1,\cdots,n$.

\end{theorem} 
\noindent 
\textit{Proof:} See supporting information (SI).\\

\vspace{0.1in}

\noindent
The system of equations, $\bm{\rho_n}\bold{(c)=0}$ in \eqref{BLML_eq2} is monotonic, i.e., $\bold{\frac{\mathrm{d}\bm{\rho_n}}{\mathrm{d}c}} > \bold{0}$,  with discontinuities at each $c_i=0$. Therefore, there are $2^n$ solutions, with each solution located in each orthant, identified by the orthant vector  $\bold{c}_0\triangleq \mathrm{sign}(\bold{c})$. Each solution corresponds to a local maximum of the likelihood function which is also its maximum value in that orthant. Hence, the global maximum always exists and can be found by finding the maximum of these $2^n$ maxima. However, it is computationally exhaustive to solve \eqref{BLML_eq2}, which entails finding the  $2^n$ solutions of $\bold{\rho_n(c)=0}$ and then comparing values of $\prod\frac{1}{c_i^2}$ for each solution. 

Therefore, to compute the BLML estimator, three approximate algorithms are proposed: a binary quadratic programming (\texttt{BLML-BQP}) algorithm, a \texttt{BLMLTrivial} algorithm, and its quicker implementation - \texttt{BLMLQuick}. Both theory and simulations show that the \texttt{BLML-BQP} algorithm is appropriate when the sample size is $n<100$ and no additional knowledge is known other than the pdf is BL. However, in cases when $n>100$ and the underlying pdf is strictly positive, the \texttt{BLMLTrivial} and \texttt{BLMLQuick} algorithms are more appropriate as they are guaranteed to yield a consistent estimate (see Theorems 4, 5 and 6) and converge at a rate ($\sim \frac{1}{n}$), which is faster than all tested KD estimates. Further, the \texttt{BLMLQuick} algorithm shows a remarkable improvement in computational speed over tested KD methods. 


\subsection{Consistency of the BLML Estimator}\label{convergence} 
Proving consistency of the BLML estimator is not trivial as it requires a solution to \eqref{BLML_eq2}. However, if $f(x)>0 \ \ \forall x$  then consistency of BLML estimator can be established. To show this, first an asymptotic solution $\bold{\bar{c}_{\infty}}$ to $\bm{\rho_n}(\bold{c)=0}$ is constructed (Theorem 3). Then, consistency is established by plugging $\bold{\bar{c}_{\infty}}$ into \eqref{BLML_eq} to show that the ISE and hence the MISE between the resulting density, $f_{\infty}(x),$ and $f(x)$ is 0  (Theorem 4). Then, it is shown that the KL-divergence between $f_{\infty}(x)$ and $f(x)$ is also $0,$ and hence $\bold{\bar{c}_{\infty}}$ is a solution to \eqref{BLML_eq2}, which makes $f_{\infty}(x)$ the BLML estimator $\hat{f}(x)$ (Theorem 5). Theorems~2-5 and their proofs are presented in SI.

\subsection{Generalization of the BLML Estimators to Joint Pdfs}

Consider the joint pdf $f(\bold{x})$, $\bold{x}\in \mathbb{R}^m,$ such that its Fourier transform $F(\bm{\omega})\triangleq \int f(\bold{x})e^{-j\bm{\omega}^T\bold{x}} d\bold{x}$ has the element-wise cut off frequencies in vector $\bm{\omega}^{true}_c\triangleq 2\pi \bold{f}_c^{true}$. Then the BLML estimator is of the following form:
\begin{equation}\label{BLMLgen_eq}
\hat{f}(x)=\left(\frac{1}{n}\sum_{i=1}^n\hat{c}_i \sinc_\bold{f_c}({\bf x-x}_i)\right)^2
\end{equation}
where, $\bold{f_c}\in \mathbb{R}^m$ is the assumed cutoff frequency, vector $\bold{x}_i's  \ \  i=1\cdots n$ are the data samples, $\sinc_\bold{f_c}(\bold{x})\triangleq \prod_{j=1}^{m}\frac{\sin(\pi f_{cj}x_j)}{\pi x_j}$ and the vector $\bold{\hat{c}}\triangleq \left[ \hat{c}_1, \cdots, \hat{c}_n \right]^T$, is given by
\begin{equation}\label{BLMLgen_eq2}
\hat{\bold{c}}=\argmax_{ \bm{\rho_n}(\bold{c)=0} }\left(\prod \frac{1}{c_i^2}\right).
\end{equation}
\noindent
Here $\rho_{ni}(\bold{c})\triangleq \sum_{j=1}^n c_js_{ij}-\frac{n}{c_i};\ s_{ij}\triangleq \sinc_\bold{f_c}({\bf x}_i-{\bf x}_j)$.

The multidimensional result can be derived in a very similar way as the one-dimensional result as described in SI.

\subsection{Computing the BLML Estimator}
 $~$ The three algorithms, \texttt{BLMLTrivial}, \texttt{BLMLQuick} and \texttt{BLML-BQP} are described next.

\paragraph{\texttt{BLMLTrivial Algorithm}.}

It is a one-step algorithm that first selects an orthant in which the global maximum may lie, and then solves $\bold{\rho_n(c)=0}$ in that orthant. As $\bold{\rho_n(c)=0}$ is monotonic, it is computationally efficient to solve in any given orthant.

As stated in Theorem 6 (see SI), the asymptotic solution of \eqref{BLML_eq2} lies in the orthant with indicator vector $c_{0i}=1\ \forall i=1,\cdots,n$ if $f(x)$ is BL and $f(x)>0 \ \  \forall \ x \in \mathbb{R}$. Therefore, the \texttt{BLMLTrivial} algorithm selects the orthant vector $c_0=\pm [1,1,\cdots,1]^T$, and then $\bold{\rho_n(c)=0}$ is solved in that orthant to compute $\bold{\hat{c}}$. It is important to note that when $f(x)$ is indeed BL and strictly positive, then the \texttt{BLMLTrivial} estimator converges to BLML estimator asymptotically.
 
Due to its simplicity, the computational complexity of the \texttt{BLMLTrivial} method is very similar to KDE with complexity $\mathcal{O}(nl)$, where $l$ is the number of points where the value of pdf is estimated \cite{Raykar2010}); but the \texttt{BLMLTrivial} method has an extra step of solving equation $\bold{\rho_n(c)=0}.$ This equation can be solved in $\mathcal{O}(n^2)$, using gradient descent or Newton algorithms. Therefore, the computational complexity of \texttt{BLMLTrivial} estimator is $\mathcal{O}(n^2+nl)$.

\paragraph{\texttt{BLMLQuick Algorithm}.}
The BL assumption of the true pdf allows for a quick implementation of the \texttt{BLMLTrivial} estimator -  ``\texttt{BLMLQuick}". For details, see SI. Briefly, \texttt{BLMLQuick} first groups the observed samples into bins of size $<\frac{0.5}{f_c}$. Then, it constructs the \texttt{BLMLTrivial} estimator of the discrete pdf (or the probability mass function, pmf) that generated the binned data. The true pmf for the binned data has infinite-bandwidth. 
Hence, under the required conditions, the \texttt{BLMLTrivial} estimate constructed using the Nyquist frequency, $2f_c,$ converges to the continuous pdf $\bar{f}(x)$, from which the pmf is obtained via sampling. $\bar{f}(x)$ can be made arbitrarily close to the true pdf $f(x)$ by choosing smaller and smaller bins. In fact, if the bin size reduces as $n^{-0.25},$ then the ISE between  $\bar{f}(x)$ and $f(x)$ is of $\mathcal{O}(1/n)$. Therefore, the MISE for \texttt{BLMLQuick} is  $\mathcal{O}(1/n)$ plus the MISE of the \texttt{BLMLTrivial} estimator. Since the MISE of the \texttt{BLMLTrivial} estimator has to be greater than $\mathcal{O}(1/n)$, the \texttt{BLMLQuick} algorithm is as efficient as the \texttt{BLMLTrivial} algorithm. Specifically, the computational complexity of \texttt{BLMLQuick} is \\$\mathcal{O}\left(n+f_c^2 n^{0.5+2/(r-1)} +f_c n^{0.25+1/(r-1)} l\right)$, where $\frac{1}{x^r}$ governs the behavior of the tail of the true pdf. 

\paragraph{\texttt{BLML-BQP Algorithm}.}

To derive the \texttt{BLML-BQP} algorithm, it is first noted that
the $2^n$ solutions of ${\rho_n(\bf c)=0}$ are equivalent to the $2^n$ local solutions of:
\begin{equation} \label{test}
\tilde{\bold{c}}=\arglocalmax_{\bold{c}^T\bold{S}\bold{c}=n^2}(\prod_i c_i^2).
\end{equation}
here $\bold{S}\in \mathbb{R}^{n\times n}$ is a matrix with $i, j$th element being $s_{ij}$. Now, if ${\bf c}_0\in \{1,-1\}^n$ is an orthant indicator vector and $\lambda \geq 0$ is such that ${(\lambda {\bf c}_0)^T{\bf S}(\lambda {\bf c}_0)}=n^2$, then \eqref{test} implies:  
\begin{equation}\label{upperbound0_eq}
\prod_i \tilde{c}_i^2 \geq \lambda^{2n} \Rightarrow \prod_i \frac{1}{\tilde{c}_i^2} \leq \frac{\bold{(c}_0^T\bold{Sc_0)}^{n}}{n^{2n}}.
\end{equation}
\noindent

Finally, the orthant where the solution of \eqref{BLML_eq2} lies is found by maximizing the upper bound $\frac{(\bold{c}_0^T\bold{Sc_0})^{n}}{n^{2n}}$ using the following BQP:
\begin{equation}
\bold{\hat{c}_0}=\argmax_{\bold{c_0} \in \{-1,1\}^n}(\bold{c}_0^T\bold{Sc_0}).
\end{equation}
BQP problems are known to be NP-hard \cite{Merz2002}, and hence a heuristic algorithm implemented in the gurobi toolbox \cite{Taylor2011} in MATLAB is used to find an approximate solution $\hat{\bold{c}}_0$ in polynomial time. Once a reasonable estimate for the orthant $\hat{\bold{c}}_0$ is obtained, ${\rho_n(\bf c)=0}$ is solved in that orthant to find an estimate for $\bold{\hat{c}}$. To further improve the estimate, the solutions to ${\rho_n(\bf c)=0}$ in all nearby orthants (Hamming distance equal to one) of the orthant $\hat{\bold{c}}_0$ are obtained and subsequently $\frac{1}{\tilde{c}_i^2}$ is evaluated in these orthants. The neighboring orthant with the largest $\frac{1}{\tilde{c}_i^2}$  is set as $\hat{\bold{c}}_0,$ and the process was repeated. This iterative process is continued until $\frac{1}{\tilde{c}_i^2}$ in all nearby orthants is no greater than  that of the current orthant. The \texttt{BLML-BQP} is computationally expensive, with complexity $\mathcal{O}(n^2+nl+BQP(n))$ where $BQP(n)$ is the computational complexity of solving BQP problem of size $n.$ Hence, the \texttt{BLML-BQP} algorithm can only be used on data samples $n<100$.

\section{Results}
A comparison of \texttt{BLMLTrivial} and \texttt{BLML-BQP} algorithms on surrogate data generated from known pdfs is presented first. Then, the performance of the \texttt{BLMLTrivial} and \texttt{BLMLQuick} algorithms is compared to several KD estimators. Finally, we show  the application results of BLML, KD and GLM methods to neuronal spiking data.

\subsection{Performance of {\small \texttt{BLMLTrivial}} versus {\small \texttt{BLML-BQP}} on Surrogate Data}
In Figure \ref{fitandmse}, \texttt{BLMLTrivial} and \texttt{BLML-BQP} estimates are presented assuming that the true pdfs are BL by $f_c=f_c^{true}$. Panels (A, C) and (B, D) use surrogate data generated from a non-strictly positive pdf $f_x=0.4\sinc^2(0.4x)$ and strictly positive pdf $f(x)=\frac{3\times 0.2}{4} (\sinc^4(0.2x)+\sinc^4(0.2x+0.1)),$ respectively. Both pdfs are BL from $(-0.4,0.4)$. In Panels A and B, the BLML estimates ($n=81$) are plotted using both algorithms, and the true pdfs are overlayed for comparison. In Panels C and D, the MISE is plotted as a function of sample size $n$ for both algorithms and both pdfs. For each $n$, data were generated 100 times to generate 100 estimates from each algorithm. The mean of the ISE was then taken over these 100 estimates to generate the MISE plots.

As expected from theory, the \texttt{BLML-BQP} algorithm works best for the non-strictly positive pdf, whereas the \texttt{BLMLTrivial} algorithm is marginally better for the strictly positive pdf. Note that as $n$ increases beyond $100,$ the \texttt{BLML-BQP} algorithm becomes computationally expensive, therefore the \texttt{BLMLTrivial} and \texttt{BLMLQuick} algorithms are used in the remainder of this paper with the assumption that the true pdf is strictly positive.

\figuremacroW{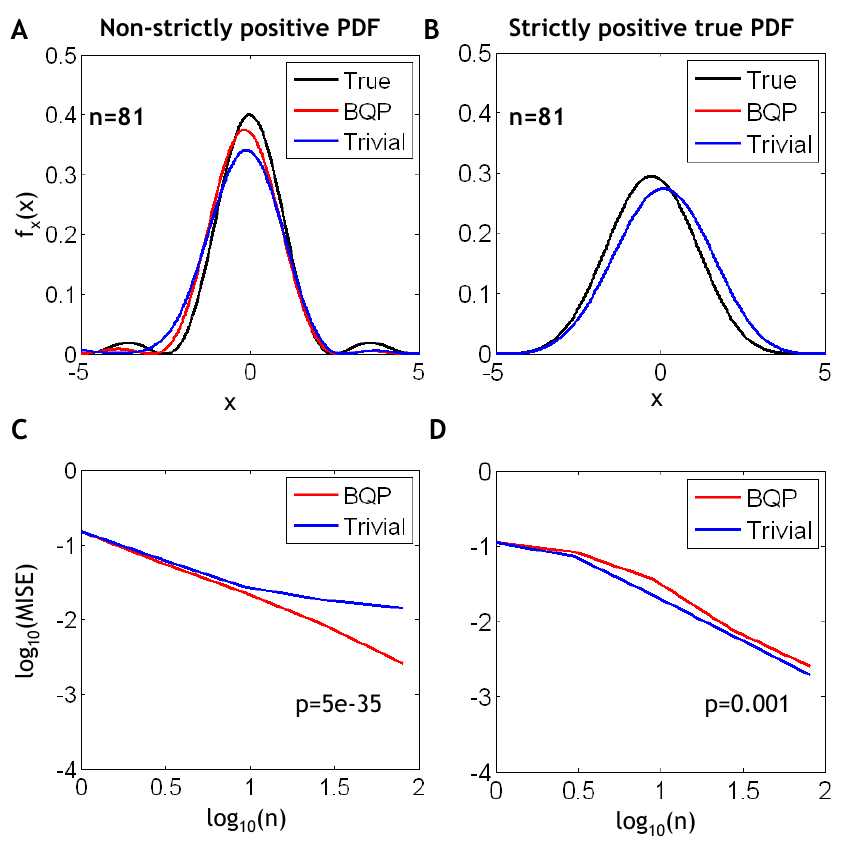}{Comparison of \texttt{BLMLTrivial} and \texttt{BLML-BQP}}{ In this plot, we show the results of \texttt{BLMLTrivial} and \texttt{BLML-BQP} algorithms using a non-strictly positive true pdf $f(x)=0.4\sinc^2(0.4x)$, (\textbf{A,C}) and a strictly positive pdf $f(x)=\frac{3\times 0.2}{4} (\sinc^4(0.2x)+\sinc^4(0.2x+0.1)),$ (\textbf{B,D}). The cut-off frequency was assumed to be  $f_c=f_c^{true}$. The $p$-values were calculated using a paired $t$-test at $n=81$.}{0.5}

\subsection{BLML and KDE on Surrogate Data}
The performance of the \texttt{BLMLTrivial} and \texttt{BLMLQuick} estimates is compared with adaptive KD estimators which are the fastest known nonparametric estimators with convergence rates of $\mathcal{O}(n^{-4/5})$, $\mathcal{O}(n^{-12/13})$ and $\mathcal{O}(n^{-1})$ for 2nd-order Gaussian (KDE2nd), 6th-order Gaussian (KDE6th) and sinc (KDEsinc) kernels, respectively \cite{Hall1987,Sheather1996}. Panels A and B of Figure \ref{gaussian} plot the MISE of the BLML estimators using the \texttt{BLMLTrivial}, \texttt{BLMLQuick}, and the adaptive KD approaches for cases in the presence of BL or non-BL pdf, respectively. In the BL case, the true pdf is strictly positive and is the same as used above, and for the infinite-band case, the true pdf is \textit{normal}. For the \texttt{BLMLTrivial}, \texttt{BLMLQuick} and sinc KD estimates, $f_c=2 f_c^{true}$ and $f_c=2$ are used for the BL and infinite-band cases, respectively. For the 2nd and 6th-order KD estimates, the optimal bandwidths ($q=\frac{0.4}{f_c} n^{-1/5}$ and $q=\frac{0.4}{f_c} n^{-1/13}$ respectively) are used. The constant $\frac{0.4}{f_c}$ ensures that MISEs are matched for $n=1$. 

It can be seen from the Figure that for both the BL and infinite-band cases, \texttt{BLMLTrivial} and \texttt{BLMLQuick} outperform KD methods. In addition, the BLML estimators seem to achieve a convergence rate that is as fast as the KDEsinc, which is known to have a convergence rate of $\mathcal{O}(n^{-1})$. Figure \ref{gaussian} C plots the MISE as function of the cut-off frequency $f_c$ for the BL pdf.  \texttt{BLMLTrivial} and \texttt{BLMLQuick} seem to be most sensitive to the correct knowledge of $f_c$, as it shows larger errors when $f_c<f_c^{true},$ which quickly dip as $f_c$ approaches $f_c^{true}$. When $f_c>f_c^{true},$ the MISE increases linearly and the BLML methods have smaller MISE as compared to KD methods. 

Finally, Figure \ref{gaussian}D plots the computational time of the BLML and KD estimators. All algorithms were implemented in MATLAB, and in-built MATLAB 2013a algorithms were used to compute the 2nd and 6th-order adaptive Gaussian KD and sinc KD estimators. The results concur with theory and illustrate that \texttt{BLMLTrivial} is slower than KD approaches for large number of observations, however, the \texttt{BLMLQuick} algorithm is remarkably quicker than all KD approaches and \texttt{BLMLTrivial} for both small and large $n$. 

\figuremacroW{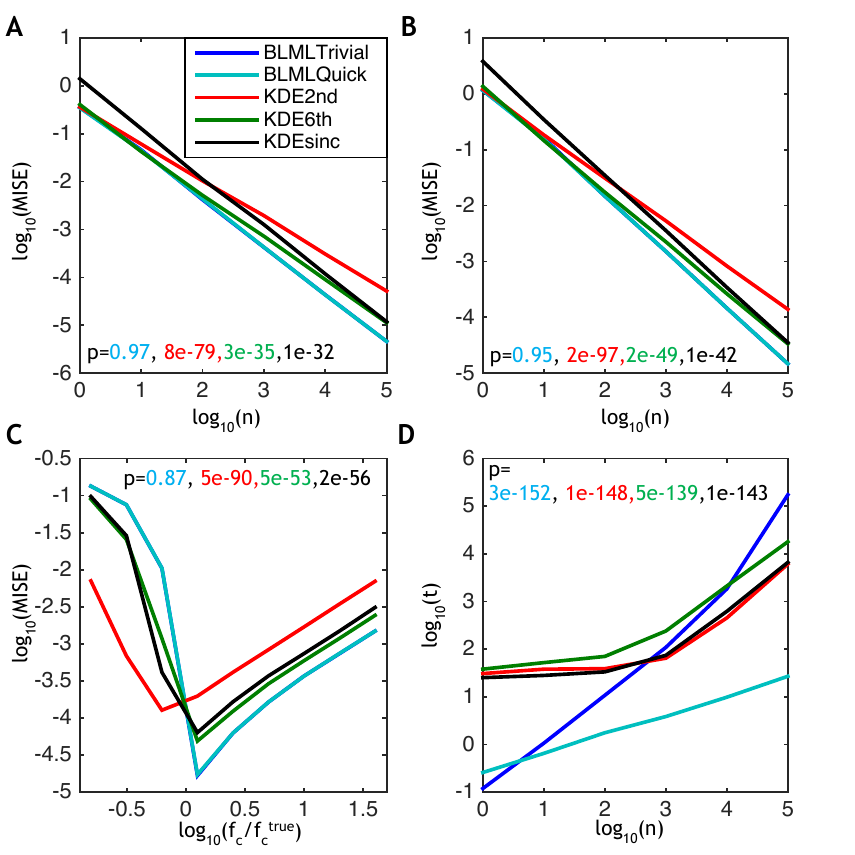}{Comparison of BLML and KD estimation}{In this plot we compare the results of the \texttt{BLMLTrivial} and \texttt{BLMLQuick} estimators to the KDE2nd, KDE6th and sinc KD estimators. Comparison of the MISE as a function of $n$ for  (\textbf{A}) a strictly positive band-limited true pdf (the one used in Figure \ref{fitandmse} B)  and  (\textbf{B}) \textit {an infinite band Gaussian normal pdf}. For the BLML estimators the cut-off frequencies are chosen as $f_c=2f_c^{true}$ for the BL true pdf and $f_c=2$ for the normal true pdf. For the KDE2nd and KDE6th, the optimal bandwidths were chosen as $q=\frac{0.4}{f_c} n^{-0.2}$ and $\frac{0.8}{f_c}n^{-1/13},$ respectively and also to match the MISE for the BLML estimator for $n=1$. For the KDEsinc, the $f_c$ was kept the same as the $f_c$ for BLML estimators. (\textbf{C}) The MISE as a function of the cutoff frequency $\frac{f_c}{f_c^{true}}$ for a BL true pdf with cut-off frequency $f_c^{true}$. $n=10^4$ was used for creating this plot. (\textbf{D}) Computation time as a function of $n$. The p-values were calculated between the \texttt{BLMLTrivial} estimator and other estimators using paired t-tests for either $\log_{10}(n)=5$ (\textbf{A,B,D}) or $\log_{10}(f_c/f_c^{true})=1.6$ (\textbf{C}) and are color coded.}{0.5} 
 
\subsection{BLML Applied to Neuronal Spiking Data}
\label{grid}

Neurons generate action potentials in response to external stimuli and intrinsic factors, including the activity of its neighbors. The sequence of action potentials over time can be abstracted as a point process, where the timing of action potentials or ``spikes" carry important information. The stochastic point process is characterized by a conditional intensity function (CIF), denoted as $\lambda$~\cite{Cox2000}. 

Here, the BLML, KD, and GLM methods are applied to estimate the CIF of a ``grid cell'' from   spike train data. In the experimental set up, the Long-Evans rat was freely foraging in an open field arena of radius of 1m for a period of 30-60 minutes. Custom microelectrode drives with variable numbers of tetrodes were implanted in the rat's medial entorhinal cortex and dorsal hippocampal CA1 area. Spikes were acquired with a sampling rate of 31.25 kHz and filter settings of 300 Hz-6 kHz.  Two infrared diodes alternating at 60 Hz were attached to the drive of each animal for position tracking. Spike sorting was accomplished using a custom manual clustering program (Xclust, M.A. Wilson). All procedures were approved by the MIT Institutional Animal Care and Use Committee.

The     spiking activity of grid cell is known to be place-modulated, whose peak firing locations define a grid-like array covering much of the 2-dimensional arena. A spike histogram of the selected cell as a function of the rat's position is shown in Figure~ \ref{ratlambda}A, which plots the $(x,y)$ coordinates of the rat's position when the cell generate spikes (red dots) and the rat's trajectory inside the arena (blue trajectory). The CIF was then estimated as a function of the rat's position (stimuli) and the neuron's spiking history (intrinsic factors):

\begin{center}
$\lambda(t|x,y,h) \triangleq \underset{\Delta t\to 0}{\lim} \frac{\Pr(\text{spike in time } \Delta t|X=x,Y=y,\mathcal{H}_t=h)}{\Delta t}$
\end{center}
where $x(t),y(t)$ is the rat's position over time inside an arena, and the vector $\mathcal{H}_t,$ consists of spiking history covariates at time $t$ as in \cite{Kass2001,Sarma2012,Sarma2010,Santaniello2012,Kahn2011}. 

Baye's rule \cite{Gelman2003} allows one to use nonparametric approaches to estimate $\lambda(\cdot)$ as follows \cite{Kloosterman2013}:
\begin{equation}
\lambda(t|x,y,h)\simeq \frac{N}{T} \frac{f(x,y,h|\text{spike in time } \Delta t)}{f(x,y,h)}
\end{equation}
where $h(t)\triangleq \log(\text{time since last spike})$, $N$ is the total number of spikes within time interval $T$, which is the total duration of the spike train observation. $f(x,y,h)$ and $f(x,y,h|\text{spike in time } \Delta t)$ are densities which are estimated using both KDE2nd (higher order kernels were too slow for estimation) and \texttt{BLMLQuick} methods. The use of the logarithm allows for a smoother dependence of $\lambda$ on $h,$ which in turn allows for capturing high frequency components in the CIF due to refractoriness (i.e., sharp decrease in $\lambda(t)$ after a spike) and bursting. 

The bandwidths and cut-off frequencies used are $q_x=q_y=n^{-0.2},q_h=0.5n^{-0.2}$ and  $f_{cx}=f_{cy}=1.4,f_{ch}=1.75$  for KDE2nd and \texttt{BLMLQuick} respectively. These bandwidths and cut-off frequencies were chosen after testing different combinations, and the frequencies and bandwidths that best fits the test data (i.e., had the lowest KS statistics, see below) for each method were used. Since the rat cannot leave the circular arena, the estimates $f(x,y,h|\text{spike in time } \Delta t)$ and $f(x,y,h)$ are normalized to integrate to 1 within the arena. The nonparametric estimates of $\lambda$ are also compared with two popular GLMs:

\begin{enumerate}
\item Gaussian GLM (GLMgauss)
\begin{small}
\begin{equation}\label{glm_eq}
\log(\lambda(x,y,\mathcal{H}))=\alpha_1 +\alpha_2 x +\alpha_3 y +\alpha_4 x^2+\alpha_5 y^2+\alpha_6 xy +\sum_{i=1}^{25}\beta_i h_i
\end{equation} 
\end{small}
\item Zernike GLM (GLMzern)
\begin{equation}\label{glmzern_eq}
\log(\lambda(x,y,\mathcal{H}))=\sum \gamma_{i,j} \chi(i,j) +\sum_{i=1}^{25}\beta_i h_i
\end{equation} 

\end{enumerate}
where $\alpha_1,\cdots,\alpha_6,\beta_i,\gamma_{i,j}$ are the parameters estimated from the data. $\mathcal{H}\triangleq [h_1,\cdots,h_{25}]^T$ where $h_i$ are the number of spikes in $(t-2i,~t-2i+2)ms$ and $\chi(i,j)$ are Zernike polynomials of 3rd order. 

The goodness-of-fit of the four estimates of $\lambda$ are computed using the time rescaling theorem and the Kolmogorov-Smirnov (KS) statistic~\cite{Brown2002}. Briefly, 80\% of the data is used to estimate $\lambda$ and then the empirical CDF of rescaled spike times is computed using the remaining 20\% test data, which should follow a uniform CDF if the estimate of $\lambda$ is accurate. The similarity between the two CDFs is quantified using the normalized KS-statistic and visualized using the KS-plot. A value of KS$>1$ indicates that the estimated $\lambda$ is too extreme ($p<0.05$) to generate the test data. The closer the normalized KS-statistic is to $0$, the better the estimate. 

Figure \ref{ratlambda} B, shows the KS-plots \cite{Brown2002} for each estimator. It is clear from the figure, that the \texttt{BLMLQuick} estimate is the only model for which the KS-plot remains inside the $95\%$ confidence bounds with KS$= 0.65$. The KS for KDE2nd, GLM Gaussian and GLM Zernike methods were $1.29$, $4.98$ and $7.47$, respectively. 

Finally, Figure \ref{ratlambda}C, plots the CIF estimates of $\lambda(t)$ using the four methods for a sample period of $400ms.$  The \texttt{BLMLQuick} method allows for a sharper and taller $\lambda(t)$, than the GLM and KDE2nd methods and it successfully captures the known behaviour of refractoriness and bursting in the neuronal activity. Although, in this instance the \texttt{BLMLQuick} method outperforms the KD and GLM methods, it is not yet clear whether this will always be the case. Several other model structures for the GLM methods must be tested, and there certainly will be receptive fields of neurons where the relationship to external covariates is more Gaussian-like and the GLM Gaussian or GLM Zernike may do better than \texttt{BLMLQuick}. However, for neurons like the grid cell, where the receptive field's dependence on external covariates does not follow any particular known profile, nonparametric methods like \texttt{BLMLQuick} or KDE methods may be more appropriate. 

\figuremacroW{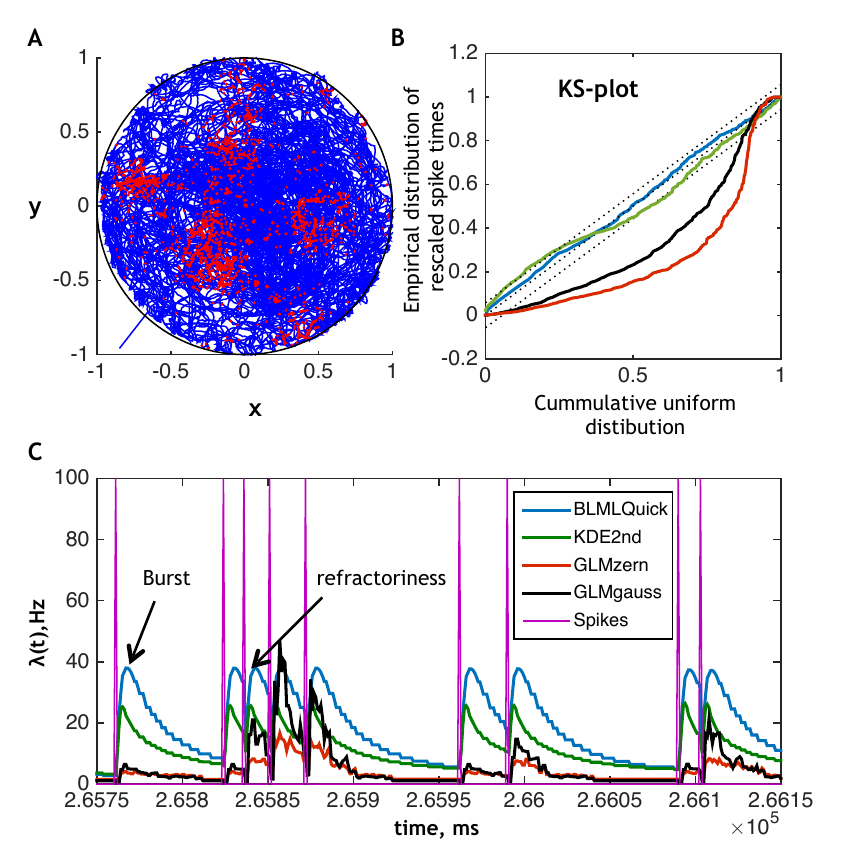}{Comparision of BL-MLE(trivial), KDE and GLM methods for  grid cell data}{ (\textbf{A}) The trajectory of rat during the duration of the experiment is shown by black line. Each dot marks (x,y) co-ordinates of rat's position when the place cell spikes. (\textbf{B}) The KS-plot for \texttt{BLMLTrivial} (blue), KDE2nd (purple), GLM Gaussian (black) and GLM Zernike (red), along with 95\% confidence intervals (dashed ines) using 20\% of test data. (\textbf{C}) Estimated $\lambda(t)$ using the four methods. All four methods show refratoriness and bursting.}{0.5} 

\section{Discussion}
In this paper, a nonparametric ML estimator for BL densities is developed and its consistency is proved. In addition, three heuristic algorithms that allow for quick computation of the BLML estimator are presented. Although these algorithms are not guaranteed to generate the BLML estimate, we show that for strictly positive pdfs, the \texttt{BLMLTrivial} and \texttt{BLMLQuick} estimates converge to the BLML estimate asymptotically. Further, \texttt{BLMLQuick} is remarkably quicker than all tested KD methods, while maintaining convergence rates of BLML estimators. Even further, using surrogate data, it is shown that both the \texttt{BLMLTrivial} and \texttt{BLMLQuick} estimators have an apparent convergence rate of $1/n$ for MISE, which is equal to that of parametric methods.  Finally, BLML is applied to spiking data, where it outperforms state-of-the-art estimation techniques used in neuroscience. 

The BLML estimators may be motivated by quantum mechanics. The function $g(x)$ in the development of BLML estimate (see SI) is analogous to the wave function \cite{Rae2008} in quantum mechanics, where the square of the absolute value of both are probability density functions. In addition, in quantum mechanics the wave function of momentum is the Fourier transform of the wave function of position. Therefore, if the momentum wave function has finite support, then the position wave function is BL and vice versa. Such occurrences are frequent in the single or double slit experiment, where one observes bandlimted ($\sinc^{2}(f_1x)$ and $\cos^2(f_2x)$ respectively) profile for the probability of finding a particle at a distance $x$ from the center. Also, in the thought experiment of a particle in a box: the wave function for position has finite support, making the momentum wave function BL. We suspect that a large number pdfs in the nature are BL because macro world phenomenon are a sum of quantum level phenomenon and pdfs at quantum level are shown to be BL (single and double slit experiments). Furthermore, the set of BL pdfs is complete, i.e. the sum of two random variables that each have a BL pdf is a random variable whose pdf is a convolution of original pdfs, and hence is BL. Therefore, if macro level phenomenon is a linear combination of different quantum level phenomenon with BL pdfs, then the macro level phenomenon will also generate a BL pdf. In fact, we see this at macro level where we observe Gaussian pdfs of various processes. The Gaussian pdf is almost BL, with cutoff frequency $f_c=10/\sigma$ ($<10^{-324}$\% of its power lies outside this band). In fact, given finite data, it is impossible to distinguish if the data is generated by a Gaussian or BL pdf.

\figuremacroW{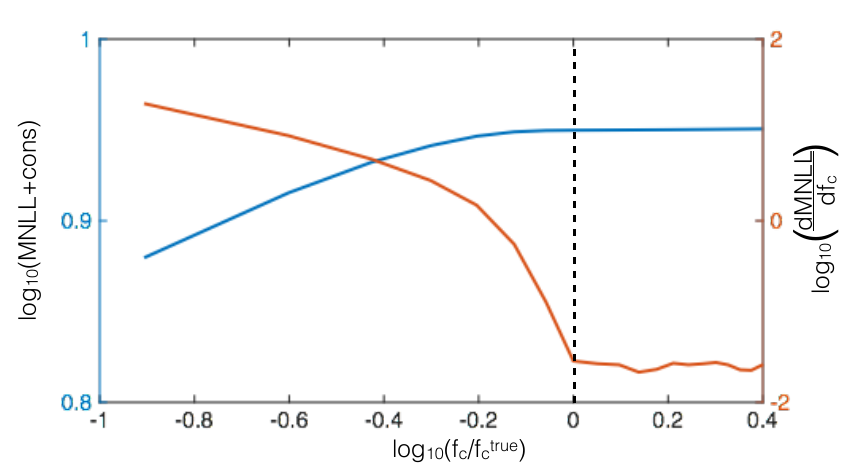}{Estimation of $f_c^{true}$}{MNLL and $\frac{d \text{MNLL}}{df_c}$ curves as a function of $f_c$. The cons is an arbitrary constant that is added to MNLL so that the logarithm of sum could exist.}{0.5}  

\subsection{Choosing A Cut-off Frequency for the BLML Estimator}
The BLML method requires selecting a cut-off frequency of the unknown pdf. One strategy for estimating the true cut-off frequency is to first fit a Gaussian pdf using the data via ML estimation. Once an estimate for standard deviation is obtained, one can estimate the cut-off frequency using the formula $f_c=1/\sigma,$ as this will allow most power of the true pdf to lie within the assumed band if the true pdf has Gaussian-like tails.

Another strategy is to increase the assumed cut-off frequency of BLML estimator as a function of the sample size. For such a strategy, the BLML estimator may converge even when the true pdf has an infinite frequency band, provided that the increase in cut-off frequency is slow enough and the cut-off frequency approaches infinity asymptotically, e.g. $\omega_c \propto \log(n)$. 

A more sophisticated strategy would be to look at the mean normalized log-likelihood (MNLL), $\mathbb{E}(-\frac{1}{n}\sum \log(\hat{c}_i^2))$ as a function of assumed cut-off frequency $f_c$. Figure~\ref{fctrue} plots MNLL (calculated using \texttt{BLMLTrivial} algorithm) is plotted for $n=200$ samples from a strictly positive true pdf $f(x)=\frac{3\times 0.2}{4} (\sinc^4(0.2x)+\sinc^4(0.2x+0.1))$ along with $\frac{d \text{MNLL}}{df_c}$. Note that $\frac{d \text{MNLL}}{df_c} \simeq \mathbb{E}(\frac{1}{n^2}\sum_{ij}\hat{c}_i\hat{c}_jo_{ij}),$ where $o_{ij}\triangleq \cos(f_c(x_i-x_j))$. We see that the MNLL rapidly increases until $f_c$ reaches $f_c^{true},$ after which the rate of increase sharply declines. There is a clear ``knee" in both MNLL and $\frac{d \text{MNLL}}{df_c}$ curves at $f_c=f_c^{true}$. Therefore, $f_c^{true}$ can be inferred from such a plot. A more complete mathematical analysis of this ``knee" is left for future work.

\subsection{Making {\small \texttt{BLMLQuick}} Even Faster}
There are several faster implementation of KD approaches such as those presented in \cite{Raykar2010, Silverman1982}. These approaches use numerical techniques to evaluate the sum of $n$ kernels over $l$ given points. Such techniques may also be incorporated while calculating the \texttt{BLMLQuick} estimator to make it even faster. Exploration of this idea will be done in a future study. 

\subsection{Asymptotic Properties of the BLML Estimator}
Although, this paper proves that the  BLML estimate is consistent, it is not clear whether it is asymptotically normal and efficient (i.e., achieving a Cramer-Rao-like bound). Studying asymptotic normality and efficiency is nontrivial for BLML estimators as one would need to first redefine asymptotic normality and extend the concepts of Fisher information and the Cramer-Rao lower bound to the nonparametric case. Therefore, we leave this to a future study. However, we postulate here that the curvature of MNLL plot might be related to Fisher information in the BLML case. In addition, although under simulations, the BLML estimator seems to achieve a convergence rate similar to its parametric counterparts ($\mathcal{O}_p(n^{-1})$) it is not proved theoretically. 


%

\begin{acknowledgments}
We want to acknowledge Dr. Mesrob Ohannessian for reading our initial manuscript, and Dr. Munther Dahleh, Dr. Rene Vidal and Ben Haeffele for valuable discussions. We thank Prof. M. A. Wilson and Dr. F. Kloosterman for providing the experimental data. S. V. Sarma was supported by the US National Science Foundation (NSF) Career Award 1055560, the Burroughs Wellcome Fund CASI Award 1007274 and NSF EFRI-M3C. Z. Chen was supported by the NSF-CRCNS Award 1307645.
\end{acknowledgments}


\newpage
\section{Supporting Information}
\subsection{Preliminaries and Formulation of the BLML Estimator}

Consider a pdf, $f(x)$, of a random variable $x \in \mathbb{R}$ with Fourier transform $F(\omega)\triangleq \int f(x) e^{-i \omega x} dx$. Let $\mathbb{U}(\omega_c)$ be the set of band-limited pdfs with frequency support in $(-\omega_c,\omega_c),$ i.e., $\omega_c\in \mathbb{R}$ is the cut-off frequency of the Fourier transform of the pdf. Then,
\begin{small}
\begin{myequation}
\mathbb{U}(\omega_c)= \left\lbrace f:\mathbb{R}\rightarrow \mathbb{R}^+ \ |  \int f(x) dx=1,  F(\omega) =0 \  \forall |\omega| > \omega_c   \right\rbrace
\end{myequation}
\end{small}
Since each element $f \in \mathbb{U}(\omega_c)$ is a pdf, it can be written as $f(x)=g^2(x)$, where $g\in \mathbb{V}(\omega_c)$ defined as:
\begin{small}
\begin{myequation}
\mathbb{V}(\omega_c)= \left\lbrace g:\mathbb{R}\rightarrow \mathbb{R} \ | \  g(x)=\sqrt{f(x)} , \  f \in \mathbb{U}(\omega_c)  \right\rbrace
\end{myequation}
\end{small}
Finally, $\mathbb{W}(\omega_c)$ can be defined as the set of all Fourier transforms of elements in  $\mathbb{V}(\omega_c)$:
\begin{small}
\begin{myequation}
\mathbb{W}(\omega_c)= \left\lbrace G:\mathbb{R}\rightarrow \mathbb{C} \ |\  G(\omega)=  \int g(x) e^{-i \omega x} dx, \   g \in \mathbb{V}(\omega_c)    \right\rbrace
\end{myequation}
\end{small}
Note that since $f(x) \in \mathbb{U}(\omega_c)$ is band limited, $g(x) \in \mathbb{V}(\omega_c)$ will also be band limited in $(\frac{-\omega_c}{2},\frac{\omega_c}{2})$. Therefore, $G(\omega)=0 \ \forall |\omega| > \frac{\omega_c}{2} \  \forall G \in \mathbb{W}(\omega_c)$.  Finally, $\mathbb{V}(\omega_c)$ and $\mathbb{W}(\omega_c)$ are Hilbert spaces with the inner product defined as $<a,b>= \int a(x)b^*(x) dx$, $<a,b>=\frac{1}{2\pi} \int a(\omega)b^*(\omega) d\omega$, respectively. The norm $||a||_2^2=<a,a>$ is defined for both spaces. Further, note that for all elements in $\mathbb{V}(\omega_c)$ and $\mathbb{W}(\omega_c)$, $||a||_2^2=<a,a>=1$.

\paragraph{The Likelihood Function for Band-limited Pdfs} 

Now consider a random variable, $x\in \mathbb{R}$, with unknown pdf $f(x) \in \mathbb{U}(\omega_c)$ and its $n$ independent realizations $x_1,x_2,\cdots,x_n$. The likelihood $L(x_1,\cdots,x_n)$ of observing $x_1,\cdots,x_n$ is then:

\begin{small}
\begin{mysubequations}\label{fhat_eq}
\begin{align}
L(x_1,\cdots,x_n)  & =  \prod_{i=1}^n f(x_i)=\prod_{i=1}^n g^2(x_i), \  g\in \mathbb{V}(\omega_c) \\
& =  \prod_{i=1}^n \left(\frac{1}{2\pi}\int G(\omega) e^{j\omega x_i}d\omega \right)^2, \ G \in \mathbb{W}(\omega_c)
\end{align}
\end{mysubequations}
\end{small}
Defining:
\begin{myequation} \label{bi_eq}
b_i(\omega) \triangleq  \left\lbrace \begin{array}{cc} e^{-j\omega x_i} & \forall  \  \omega \in (-\frac{\omega_c}{2}, \frac{\omega_c}{2}) \\
0 &  o.w. \end{array} \right\rbrace
\end{myequation}
gives:
\begin{myequation}\label{likelihood_eq}
L(x_1,\cdots,x_n) =\prod_{i=1}^n \left( < G(\omega), b_i(\omega)> \right)^2 \triangleq L[G] .
\end{myequation}

Further, consider $\hat{G}(\omega)$ which maximizes the likelihood function:

\begin{myequation}\label{max_prb}
\hat{G}=\argmax_{G\in \mathbb{W}(\omega_c)}(L[G]).
\end{myequation}

\noindent
Then the BLML estimator is:

\begin{myequation}\label{fx_eq}
\hat{f}(x)=\left(\frac{1}{2\pi}\int \hat{G}(\omega) e^{j\omega x}d\omega \right)^2.
\end{myequation}

\section{Proof of Theorem 1}
%

\noindent
\textit{Proof:} Because of \eqref{bi_eq},  \eqref{max_prb} is equivalent to

\begin{myequation} \label{eq_prb}
\hat{G}(\omega)=\argmax_{G:\mathbb{R} \rightarrow \mathbb{C},||G||_2^2=1 }(L[G]).
\end{myequation}

Note that Parseval's equality \cite{Hazewinkel2001} is applied to get the constraint $||G||_2^2=1$. Now, the Lagrange multiplier \cite{Bertsekas1999} is used to convert \eqref{eq_prb} into the following unconstrained problem: 

\begin{myequation}
\hat{G}(\omega)=\argmax_{G:\mathbb{R}\rightarrow \mathbb{C}}\left(L[G]+\lambda \left(1-||G||_2^2\right)\right).
\end{myequation}

$\hat{G}(\omega)$ can be computed by differentiating the above equation with respect to $G$ using calculus of variations \cite{Gelfand2000} and equating it to zero. This gives:

\begin{small}
\begin{mysubequations}\label{resultsa_eq}
\begin{align}
\hat{G}(\omega) &=\frac{1}{n}\sum_{i=1}^n c_i b_i(\omega) \\
c_i & = \frac{n}{\lambda}\prod_{j\neq i}  \left( < \hat{G}(\omega),b_j(\omega)> \right)^2 <\hat{G}(\omega),b_i(\omega)> \nonumber \\
 &~~~~~~~~~~~~~~~~~~~~~~~~~~~~~~~~~~~~~\text{for} \ \  i=1 \cdots n 
\end{align}
\end{mysubequations}
\end{small}

To solve for $c_i,$ the value of $\hat{G}$ is substituted back from \eqref{resultsa_eq}a into \eqref{resultsa_eq}b and both sides are multiplied by $<\hat{G}(\omega), b_i(\omega)>$ to get:
\begin{small}
\begin{mysubequations}\label{results0_eq}
\begin{align}
c_i & \sum_{j=1}^n c_j < b_j(\omega), b_i(\omega)> =n^2 k \  \text{for} \  i=1 \cdots n, \\
k & \triangleq \frac{1}{n^{2n} \lambda}\left(\prod_{j=1}^{n}  \left( \sum_{i=1}^nc_i< b_i(\omega),b_j(\omega)>\right)^2\right)\\
&=\frac{1}{n^{2n} \lambda}\left(\prod_{j=1}^{n}  \left( \sum_{i=1}^nc_i s_{ij}\right)^2\right)
\end{align}
\end{mysubequations}
\end{small}

\noindent
To go from \eqref{results0_eq}b to \eqref{results0_eq}c, observe that $ < b_i(\omega), b_j(\omega)> \ =\frac{\sin(\pi f_c(x_i-x_j))}{\pi(x_i-x_j)}=s_{ij}$  (here $f_c = \frac{ \omega_c}{2\pi}$). Now by defining,

\begin{small}
\begin{myequation}\label{notation_eq}
\bold{S}\triangleq \left[ \begin{array}{ccc} s_{11} & \cdots  & s_{1n} \\ \vdots & \ddots & \vdots \\ s_{n1} & \cdots & s_{nn} \end{array} \right],
\end{myequation}
\end{small}

\noindent
and using \eqref{resultsa_eq}a and the constraint $||\hat{G}(\omega)||_2^2=1$, one can show that $\bold{c}^T\bold{Sc}=n^2$. Also, summing up all $n$ constraints in \eqref{results0_eq}a gives $\bold{c}^T\bold{Sc}=n^3k$, hence $k=1/n$. Now, substituting the value of $k$ into \eqref{results0_eq}a and rearranging terms gives the following $n$ constraints: 

\begin{small}
\begin{myequation}\label{results_eq}
\frac{1}{n}\sum_{j=1}^{n}c_js_{ij}-\frac{1}{c_i}=\rho_{ni}(\bold{c})=0 \ \ \text{for} \ \  i=1 \cdots n.
\end{myequation}
\end{small}

As mentioned in the main text, the above system of equations ($\bm{\rho_n}\bold{(c)=0}$) is monotonic, i.e., $\bold{\frac{\mathrm{d}\bm{\rho_n}}{\mathrm{d}c}} > \bold{0}$, but with discontinuities at each $c_i=0$. Therefore, there are $2^n$ solutions, with each solution located in each orthant, identified by the orthant vector  $\bold{c}_0\triangleq sign(\bold{c})$. Each of these solutions can be found efficiently by choosing a starting point in a given orthant and applying numerical methods from convex optimization theory to solve for \eqref{results_eq}. Thus, each of these $2^n$ solutions corresponds to a local maximum of the likelihood functional $L[G]$. The global maximum of $L[G]$ can then be found by evaluating the likelihood for each solution $\bold{c}=[c_1,\cdots,c_n]^T$ of \eqref{results_eq}. The likelihood value at each local maximum can be computed efficiently by using the following expression: 

\begin{small}
\begin{myequation}\label{likelihoodvalue_eq}
L(\bold{c})=\prod_i\left(\frac{1}{n}\sum_j c_j s_{ij}\right)^2=\prod_i \frac{1}{c_i^2}.
\end{myequation}
\end{small}

This expression is derived by substituting \eqref{resultsa_eq}a into \eqref{likelihood_eq} and then substituting \eqref{results_eq} into the result. Now the global maximum $\bold{\hat{c}}$  can be found by solving \eqref{BLML_eq2}. Once the global maximum $\bold{\hat{c}}$ is computed, we can put together \eqref{bi_eq},\eqref{fx_eq} and \eqref{resultsa_eq}a to write our solution as \eqref{BLML_eq}. Hence proved.\\

\subsection{Consistency of the BLML Estimator}

\subsubsection{Bounds on Bandlimited PDF}
In this section the following theorem is first stated and proved.
\vspace{0.1in}

\noindent
\begin{theorem}\label{maxval_thm}
For all $f\in \mathbb{U}(\omega_c) \ f(x)\leq \frac{\omega_c}{2\pi}\ \forall x\in\mathbb{R}$. 
\end{theorem}

\textit{Proof}: Above theorem can be proven by finding:

\begin{small}
\begin{myequation}
y=\max_{f\in \mathbb{U}(\omega_c)} \max_{x\in\mathbb{R}} f(x).
\end{myequation}
\end{small}
Because a shift in the pdf domain (e.g. $f(x-\mu)$) does not change the magnitude or bandwidth of $F(\omega)$,
without loss of generality one can assume that $\max_{x\in\mathbb{R}}f(x) = f(0)$ and write the above equation as
\begin{small} 
\begin{mysubequations}\label{maxfx_eq}
\begin{align}
y&=\max_{f\in \mathbb{U}(\omega_c)}\left(f(0)\right)\\
&=\max_{g\in \mathbb{V}(\omega_c)}((g^2(0))\\
&=\max_{G\in\mathbb{W}(\omega_c)}\left(\left(\int_{-\omega_c}^{\omega_c} G(\omega) \mathrm{d}\omega\right)^2\right)\\
&=\max_{||G||^2=1}\left(\left(\int_{-\infty}^{\infty} G(\omega)b(\omega) \mathrm{d}\omega\right)^2\right)\\
&=\max\left(\left(\int G(\omega)b(\omega) \mathrm{d}\omega\right)^2 +\lambda (||G||^2-1)\right)
\end{align}
\end{mysubequations}
\end{small}
Here $b(\omega)=1 \iff\  |\omega|<\pi f_c$ and is $0$ otherwise. Now by differentiating \eqref{maxfx_eq}e and subsequently setting the result equal to $0$, gives $G^*(\omega)=\frac{b(\omega)}{\sqrt{f_c}}$. Therefore $g^*(x)=\frac{\sin(\pi f_cx)}{\pi \sqrt{f_c} x}$, which gives $y=f_c=\frac{\omega_c}{2\pi}$.

\vspace{0.1in}
\textbf{Corollary}: By the definition of $\mathbb{V}(\omega_c)$, one can apply Theorem \ref{maxval_thm} and show that for  all $g \in \mathbb{V}(\omega_c), \ g(x) \leq \sqrt{\frac{\omega_c}{2\pi}}$.\\


\subsubsection{Sequence $\bar{c}_{nj}$:}
 Now a sequence $\bar{c}_{nj}$ is defined and some of its properties are stated and proved. These properties will be used to prove Theorems \ref{limitsol_thm} and \ref{convergence_thm} below.

\begin{small}
\begin{mysubequations}\label{cj_eq}
\begin{align}
\bar{c}_{nj}&\triangleq \frac{ng(x_j)}{2f_c} \left( \sqrt{1+\frac{4}{n}\frac{f_c}{g^2(x_j)}}-1\right) \ \forall 1\leq j \leq n
\end{align}
\end{mysubequations}
\end{small}

\subsubsection{Properties of $\bar{c}_{nj}$} 
$\bar{c}_{nj}$ has following properties:

\begin{small}
\begin{mysubequations}\label{cjproperty_eq}
\begin{align}
(P1)&\  \frac{1}{\bar{c}_{nj}}-\frac{\bar{c}_{nj}f_c}{n}=g(x_j)\\
(P2)&\  \bar{c}_{nj}=\frac{1}{g(x_j)}\left( 1 +\mathcal{O}\left(\frac{1}{ng^2(x_j)}\right) \right)\  \ \text{for} \ \  ng^2(x_j)>f_c \\
(P3)&\  \bar{c}_{nj}^2=\frac{n}{f_c}(1-\bar{c}_{nj}g(x_j))\\
(P4)&\  \sqrt{\frac{3/2-\sqrt{5}/2}{f_c}}\leq |\bar{c}_{nj}|\leq\sqrt{\frac{n}{f_c}} \\
(P5)&\  0\leq 1-\bar{c}_{nj}g(x_j)\leq 1 \\
(P6)&\ 1-\frac{1}{n}\sum_{j=1}^n \bar{c}_{nj}g(x_j) < \mathcal{O}_{a.s.}\left(\frac{1}{\sqrt{n}}\right) \  if \ g(x)>0 \forall x \\
(P7)&\  \frac{1}{n}\sum_{j\neq i}(s_{ij}\bar{c}_{nj})=\frac{1}{n}\sum_j s_{ij}\bar{c}_{nj}-\mathcal{O}\left(\frac{1}{\sqrt{n}}\right) \nonumber \\ &\ \ =g(x_i)+\epsilon_{ni}  \overset{a.s.}\to g(x_i) \text{ simultaneously } \forall i \  if \ g(x)>0 \forall x\\
(P8)&\ \bar{c}_{\infty j}\triangleq \lim_{n\to\infty} \bar{c}_{nj}\geq \bar{c}_{nj} \ \forall n 
\end{align}
\end{mysubequations}
\end{small}

\subsubsection{Proofs for Properties of $\bar{c}_{nj}$}
(P1) can be proved by direct substitution of $\bar{c}_{nj}$ into left hand side (LHS). (P2) can be derived through binomial expansion of $\bar{c}_{nj}$. (P3) can again be proved by substituting $\bar{c}_{nj}$ and showing LHS=RHS. (P4) and (P5) can be proved by using the fact that both $\bar{c}_{nj}^2$ and $\bar{c}_{nj} g(x_j)$ are monotonic in $g^2(x_j)$ since $\frac{d \bar{c}_{nj}^2}{d g^2(x_j)}<0$ and $\frac{d \bar{c}_{nj} g(x_j)}{d g^2(x_j)}>0$. Therefore, the minimum and maximum values of $|c_j|$ and $c_jg(x_j)$, can be found in by plugging in the minimum and maximum values of $g^2(x_j)$ (note $0 \leq g^2(x_j) \leq f_c$, from Thm \ref{maxval_thm} ). 

(P6) is proved by using Kolmogorov's sufficient criterion \cite{Kaboyashi2012} for almost sure convergence of sample mean. Clearly, from (P5) $E(\bar{c}_{nj}^2g^2(x_j)) < \infty$ which establish almost sure convergence. Now, let $\beta\triangleq \frac{1}{n}\sum \bar{c}_{nj}g(x_j)$. Then multiplying each side of $n$ equations in (P1) by $\frac{1}{g(x_j)}$, respectively, adding them and the normalizing the sum by $\frac{1}{n}$ gives:
\begin{mysubequations}\label{matrix_eq}
\begin{align}
\frac{1}{n}\sum\frac{1}{\bar{c}_{nj}g(x_j)}& =1+\frac{1}{n}\sum \frac{\bar{c}_{nj}f_c}{ng(x_j)} \\
\Rightarrow \frac{1}{\beta} & \leq 1+b_n\\
\Rightarrow \beta \geq & \frac{1}{1+b_n}
\end{align}
\end{mysubequations}

Above $b_n\triangleq \sum_j\frac{f_c\bar{c}_{nj}}{n^2g(x_j)}$. To go from \eqref{matrix_eq}a to \eqref{matrix_eq}b, the result $\frac{1}{n}\sum \frac{1}{\bar{c}_{nj}g(x_j)}\geq \frac{n}{\sum \bar{c}_{nj}g(x_j)}=\frac{1}{\beta}$ (Arithmetic Mean $\geq$ Harmonic Mean) is used. Now it can be shown that $b_n\leq\mathcal{O}_{a.s}\left(\frac{1}{\sqrt{n}}\right)$, as following:
\begin{mysubequations}\label{bn_eq}
\begin{align}
b_n=& \sum_i\frac{f_c\bar{c}_{ni}}{n^2g(x_i)}\\
\leq &\frac{\sqrt{f_c}}{n\sqrt{n}}\sum_i \frac{1}{g(x_i)} \\
\overset{a.s}{\to}&\sqrt{\frac{f_c}{n}}E\left(\frac{1}{g(x_i)}\right)\\
=& \mathcal{O}_{a.s}\left(\frac{1}{\sqrt{n}}\right)
\end{align}
\end{mysubequations}
To go from \eqref{bn_eq}a to \eqref{bn_eq}b (P4) and $g(x)>0$ are used. To go from \eqref{bn_eq}c to \eqref{bn_eq}d, $E\left(\frac{1}{g(x_i)}\right)=\int g(x_i) d x_i$ is used, which has to be bounded as $g^2(x)$ is pdf and bandlimited (due to Plancherel). Finally the fact that  the  sample mean of positive numbers, if converges, converges almost surely gives \eqref{bn_eq}d. Combining \eqref{bn_eq}d and \eqref{matrix_eq}c gives:
\begin{myequation}\label{beta_eq}
\beta\geq 1-\mathcal{O}_{a.s}\left(\frac{1}{\sqrt{n}}\right)
\end{myequation}

substituting $\beta$ in LHS of (P6) proves it. \\

To prove (P7) we first establish almost sure convergence of each eaquation seprately by using Kolmogorov's sufficient criterion \cite{Kaboyashi2012}.  By Kolmogrov's sufficient criterion:

\begin{small}
\begin{mysubequations}\label{p6_eq}
\begin{align}
&\frac{1}{n}\sum_{j\neq i} s_{ij}\bar{c}_{nj} =\frac{1}{n}\sum_j s_{ij}\bar{c}_{nj}-\mathcal{O}\left(\frac{1}{\sqrt{n}}\right)\\
&\overset{a.s.}\to E_j(s_{ij}\bar{c}_{nj})\ \ if \ E_j(\bar{c}_{nj}^2s_{ij}^2) < \infty
\end{align}
\end{mysubequations}
\end{small}

Thus, now we compute $E_j(\bar{c}_{nj} s_{ij})$ and $E_j(\bar{c}_{nj}^2 s_{ij}^2)$ as follows:

\begin{small}
\begin{mysubequations}\label{expcs_eq}
\begin{align}
&|E_j(\bar{c}_{nj} s_{ij}) -g(x_i)|\nonumber \\
&\ = \left| \int \bar{c}_{nj} s_{ij} g^2(x_j) \mathrm{d}x_j -g(x_i) \right|\\
&\ = \left|\int_{ng^2(x)\geq f_c} s_{ij}g(x_j) +\mathcal{O}\left(\frac{1}{n}\right)\frac{s_{ij}}{g(x_j)} \mathrm{d}x_j \right. \nonumber \\
&\ \ \ \ \left. +\int_{ng^2(x)<f_c} \bar{c}_{nj} s_{ij}g^2(x_j) \mathrm{d}x_j -g(x_i)\right|\\
&\ = \left|\int s_{ij} g(x_j) \mathrm{d}x_j -  \int_{ng^2(x)< f_c} (1-\bar{c}_{nj}g(x_j))s_{ij}g(x_j) \mathrm{d}x_j \right. \nonumber \\
&\ \ \ \ \left.+ \int_{ng^2(x)\geq f_c} \mathcal{O}\left(\frac{1}{n} \right) \frac{s_{ij}}{g(x_j)} \mathrm{d}x_j -g(x_i)\right| \\
&\ \leq \int_{ng^2(x)< f_c} |s_{ij}g(x_j)| \mathrm{d}x_j + \mathcal{O}\left(\frac{1}{n} \right)\int_{ng^2(x)\geq f_c} \left|\frac{s_{ij}}{g(x_j)}\right| \mathrm{d}x_j 
\end{align}
\end{mysubequations}
\end{small}

\noindent
To go from \eqref{expcs_eq}c to \eqref{expcs_eq}d, the facts that $\int s_{ij}g(x_j) \mathrm{d}x_j=g(x_i)$ for any $g\in \mathbb{V}(\omega_c)$ and (P5) are used. Now define   

\begin{center}
$\varepsilon_{n}(x_i) \triangleq \mathcal{O}\left(\frac{1}{n} \right)\int_{ng^2(x)\geq f_c} \left|\frac{s_{ij}}{g(x_j)}\right| \mathrm{d}x_j +\int_{ng^2(x)< f_c} |s_{ij}g(x_j)| dx_j.$ 
\end{center}

Then it is shown, 
\begin{small}
\begin{mysubequations}\label{epsiloncon_eq2}
\begin{align}
&|E_j(\bar{c}_{nj} s_{ij}) -g(x_i)| \leq \varepsilon_{n}(x_i)\to 0 \ \ \ \text{ uniformly} \ if \ g(x)>0
\end{align}
\end{mysubequations}
\end{small}

by first noting that 
\begin{align*}
\int_{ng^2(x)\geq f_c} \left|\frac{s_{ij}}{g(x_j)}\right| \mathrm{d}x_j\leq \sqrt{\frac{n}{f_c}}\int_{ng^2(x)\geq f_c} \left|s_{ij}\right| \mathrm{d}x_j,
\end{align*}
and that the length of limit of integration has to be less than $\frac{n}{f_c}$ as $g^2(x)$ has to integrate to 1. This makes $\int_{ng^2(x)\geq f_c} \left|s_{ij}\right| \mathrm{d}x_j\leq \mathcal{O}(\log(n))$ and hence 
\begin{align}
\mathcal{O}\left(\frac{1}{n}\right)\int_{ng^2(x)\geq f_c} \left|\frac{s_{ij}}{g(x_j)}\right| \mathrm{d}x_j\leq  \mathcal{O}\left(\frac{\log(n)}{\sqrt{n}}\right)\to 0 \text{ uniformly.} \nonumber
\end{align}

 Then, $\int_{ng^2(x)< f_c} |s_{ij}g(x_j)| dx_j < f_c\int_{ng^2(x)< f_c} g(x_j) dx_j$ $ \ if\ g(x)$  $> 0$ is also shown to go to zero uniformly, by first considering

\begin{myequation}
\zeta_n(x_j)\triangleq \left\lbrace \begin{array}{cc}
g(x_j)  & \text{if} \ \  g^2(x_j)\geq \frac{f_c}{n} \\ 0 & o.w. 
\end{array} \right.
\end{myequation}

\noindent
The sequence $\zeta_n(x_j)$ is nondecreasing under the condition $g^2(x)>0$ \& $g^2(x)\in \mathbb{U}(\omega_c)$ , i.e $\zeta_{n+1}(x_j)\geq\zeta_{n}(x_j) \ \forall \  x_j,$ and the $\lim_{n\to\infty} \zeta_n(x_j)= g(x_j)$. Therefore, by the monotone convergence theorem, $\lim_{n\to\infty} \int\zeta_n(x_j) dx_j =\int_{-\infty}^\infty g(x_j) dx_j $. This limit converges due to Plancherel. Now, by definition of $\zeta_n(x_j)$,  

\begin{mysubequations}
\begin{align}
\lim_{n\to\infty} & \int_{ng^2(x)< f_c} |s_{ij}g(x_j)| dx_j \leq \nonumber \\
& f_c\int_{-\infty}^\infty g(x_j) dx_j -\lim_{n\to\infty}f_c\int\zeta_n(x_j) dx_j \to 0 \text{ uniformly}. 
\end{align}
\end{mysubequations}

Therefore $\varepsilon_{n}(x_i)\to 0\ \text{uniformly }\forall x_i$ which is equivalent to saying $\max_x{\varepsilon(x)}\to 0$. A weaker but more informative proof for going to step \eqref{expcs_eq}e to \eqref{expcs_eq}d can be obtained by assuming a tail behaviour of $\frac{1}{|x|^r}$ for $g^2(x)$ and showing the step holds for all $r>1$, this gives $\varepsilon_{n}(x_i)=\mathcal{O}\left(\frac{1}{\sqrt{n}}\right) \ \forall x_i$. Now it is shown: 

\begin{small}
\begin{mysubequations}\label{varcs_eq}
\begin{align}
E_j(\bar{c}_{nj}^2 s_{ij}^2)&=\int \bar{c}_{nj}^2 s^2_{ij} g^2(x_j) dx_j \\
& \leq \int s^2_{ij}  dx_j =f_c <\infty \forall x_i
\end{align}
\end{mysubequations}
\end{small}

To go from \eqref{varcs_eq}a to \eqref{varcs_eq}b, (P5) and the equality $\int s^2_{ij}  dx_j=f_c$ are invoked. Finally, substituting \eqref{expcs_eq}f and \eqref{varcs_eq}b into \eqref{p6_eq}b proves that each equation go to zero almost surely but seperately. More precisely, untill now only it has been shown that there exists sets of events $E_1, E_2,\cdots, E_n$ where each set $E_i\triangleq \{s: \lim_{n\to\infty} \rho_{ni}(\bar{c}(s))=0\}$ and $P(E_i)=1$. However to establish simultaniety of convergence we need to show $P(\cap_i^\infty E_i)=1$.\\

For this, the almost sure convergence of the following $L^2$ norm:
 
\begin{myequation}\label{l2norm_eq}
\int\left(\frac{1}{n}\sum \bar{c}_{nj}s(x-x_j) -g(x)\right)^2 \mathrm{d}x \overset{a.s.}\to 0 \ \ \text{ if } g(x)>0 
\end{myequation}

is established in next section. This implies that $\frac{1}{n}\sum \bar{c}_{nj}s(x-x_j) \overset{a.s.} \to g(x)$ uniformly due to bandlimitedness of functions \cite{Protzmann2001}. This in turn implies that eqns $\frac{1}{n}\sum_j \bar{c}_{nj}s(x_i-x_j) \overset{a.s.} \to g(x_i)$ simultaneously for all $x_i$ and hence prove (P7).

(P8) can be proved easily by showing that $\frac{d\bar{c}_{nj}}{dn}>0\  \forall n$.

\subsubsection{Proof for \eqref{l2norm_eq}}
To establish  convergence of $L^2$ norm consider:

\begin{small}
\begin{mysubequations}\label{simul_eq}
\begin{align}
\int& \left(\frac{1}{n}\sum\bar{c}_{nj}s(x-x_j) -g(x)\right)^2 \mathrm{d}x \nonumber \\ 
&= \int \frac{1}{n^2}\sum_{ij}\bar{c}_{ni}\bar{c}_{nj}s(x-x_i)s(x-x_j) + g^2(x) \nonumber \\
& - \frac{2}{n}\sum_j \bar{c}_{nj}s(x-x_j)g(x) \mathrm{d}x \\
&= \frac{1}{n^2}\sum_{ij}\bar{c}_{ni}\bar{c}_{nj}s_{ij} +1 - \frac{2}{n}\sum_j \bar{c}_{nj}g(x_j)\\
&= \frac{1}{n^2}\sum_{i\neq j}\bar{c}_{ni}\bar{c}_{nj}s_{ij} + \frac{1}{n^2} \sum_i s_{ii}\bar{c}_{ni}^2 +1 - \frac{2}{n}\sum_j \bar{c}_{nj}g(x_j)\\
&= \frac{1}{n^2}\sum_{i\neq j}\bar{c}_{ni}\bar{c}_{nj}s_{ij} + \frac{1}{n} \sum_i (1-\bar{c}_{ni}g(x_i)) +1 - \frac{2}{n}\sum_j \bar{c}_{nj}g(x_j)\\
&\overset{a.s.} \to E(\bar{c}_{ni}\bar{c}_{nj}s_{ij}) -1
\end{align}
\end{mysubequations}
\end{small}

To go from \eqref{simul_eq}c to \eqref{simul_eq}d to \eqref{simul_eq}d P3 and P6. For going to \eqref{simul_eq}e the almost sure convergence proof is established in  next section\ref{sec:asc}.  

Now, $E(\bar{c}_{ni}\bar{c}_{nj}s_{ij})$ is calculated as:
\begin{mysubequations}\label{l2_eq}
\begin{align}
E(\bar{c}_{ni}\bar{c}_{nj}s_{ij})&=\int\bar{c}_{ni}\bar{c}_{nj}s_{ij} g^2(x_i)g^2(x_j)  \mathrm{d}x_i \mathrm{d}x_j \\ 
&= \int \bar{c}_{ni} g^2(x_i) (g(x_i)+\varepsilon_n(x_i)) \mathrm{d}x_i \\
&= \int \bar{c}_{ni} g^3(x_i) \mathrm{d}x_i + \int \bar{c}_{ni} g^2(x_i) \varepsilon_n(x_i) \mathrm{d}x_i \\
&= 1 +\mathcal{O}\left(\frac{1}{\sqrt{n}}\right)  + \max_{x_i}(\varepsilon_n(x_i)) \int |g(x_i)| \mathrm{d}x_i\\
&\to 1\ \ \ \ \ \ \ \ \ \ \ \ \ \ \ \ \ \  \text{ if } g(x)>0
\end{align}
\end{mysubequations}

To go from \eqref{l2_eq}a to \eqref{l2_eq}b \eqref{epsiloncon_eq2} is used.  To go from \eqref{l2_eq}c to \eqref{l2_eq}d (P6) and (P5) are used. To go from \eqref{l2_eq}d to \eqref{l2_eq}e uniform convergence of $\varepsilon_n(x)$ and $\int g(x) < \infty$ (due to plancheral) are used.  Now, combining \eqref{l2_eq} e and \eqref{simul_eq}e establishes \eqref{l2norm_eq} and subsequently almost simultaneous convergence.

\subsubsection{Proof for Almost Sure Convergence of $\frac{1}{n^2}\sum_i\bar{c}_i\bar{c}_j s_{ij}$}\label{sec:asc}
Let $S_n\triangleq \frac{1}{n^2}\sum_{i\neq j}\bar{c}_{ni}\bar{c}_{nj}s_{ij}$,  then:

\begin{mysubequations}\label{l2var_eq}
\begin{align}
Var\left(S_n\right)&= \frac{4n(n-1)(n-2)}{n^4} E\left(\bar{c}_{ni}\bar{c}_{nj}^2\bar{c}_{nl}s_{ij}s_{jm}\right) \nonumber\\
&+ \frac{2n(n-1)}{n^4} E\left(\bar{c}_{ni}^2\bar{c}_{nj}^2s_{ij}^2\right) \\ 
&-\frac{2n(n-2)(2n-3)}{n^4}E\left(\bar{c}_{ni}\bar{c}_{nj}\bar{c}_{nl}\bar{c}_{nm}s_{ij}s_{lm}\right) \nonumber \\ 
& \leq \frac{4n(n-1)(n-2)f_c}{n^4} \left(\int g(x_i) \mathrm{d}x_i\right)^2 \nonumber \\
&+ \frac{2n(n-1)}{f_cn^3} E\left(\bar{c}_{ni}^2(1-\bar{c}_{nj}g(x_j))s_{ij}^2\right) \nonumber \\
&+\frac{2n(n-2)(2n-3)}{n^4}E\left(\bar{c}_{ni}\bar{c}_{nj}|s_{ij}|\right)^2  \\
&= \mathcal{O}\left(\frac{1}{n}\right)
\end{align}
\end{mysubequations}  

To go from \eqref{l2var_eq}a to \eqref{l2var_eq}b $\int |s_{ij} s_{jm}|<f_c$ (cauchy-schwartz inequality), P5, P3 are used. To go from  \eqref{l2var_eq}b to  \eqref{l2var_eq}c $\int g(x) < \infty$ (due to plancheral), P5 and $\int |s_{ij} g(x_i)|<\sqrt{f_c}$ (cauchy-schwartz inequality)  are used.  

Now, by chebyshev inequality $\Pr(|S_{n^2}-\mu|>\epsilon)<\mathcal{O}\left(\frac{1}{n^2}\right)$, here $\mu= \lim_{n\to\infty}E(S_n)$. Hence, $\sum_{n=1}^\infty\Pr(|S_{n^2}-\mu|>\epsilon)<\infty$, therefore by Borel-Cantelli lemma, $S_{n^2}\overset{a.s.}\to \mu$. Now to show $S_n\overset{a.s.}\to\mu$, divide $S_n$ into two parts $A_n\triangleq \frac{1}{n^2}\sum_{i\neq j}\bar{c}_{ni}\bar{c}_{nj}s_{ij} I(s_{ij})$ where $I(s_{ij})$ is indicator function which if $1$ is $s_{ij}\geq 0$ and $0$ otherwise (note that $\bar{c}_{ni}>0\ \forall i$ due to the assumption $g(x)>0$), and $B_n\triangleq S_n-A_n$. Now, 
\begin{mysubequations}\label{l3var_eq}
\begin{align}
Var\left(A_n\right) &< \frac{4n(n-1)(n-2)}{n^4} E\left(\bar{c}_{ni}\bar{c}_{nj}^2\bar{c}_{nl}|s_{ij}||s_{jm}|\right) \nonumber\\
&+ \frac{2n(n-1)}{n^4} E\left(\bar{c}_{ni}^2\bar{c}_{nj}^2s_{ij}^2\right) \\ 
&-\frac{2n(n-2)(2n-3)}{n^4}E\left(\bar{c}_{ni}\bar{c}_{nj}\bar{c}_{nl}\bar{c}_{nm}|s_{ij}||s_{lm}|\right) \nonumber \\ 
& \leq \frac{4n(n-1)(n-2)f_c}{n^4} \left(\int g(x_i) \mathrm{d}x_i\right)^2 \nonumber \\
&+ \frac{2n(n-1)}{f_cn^3} E\left(\bar{c}_{ni}^2(1-\bar{c}_{nj}g(x_j))|s_{ij}|^2\right) \nonumber \\
&+\frac{2n(n-2)(2n-3)}{n^4}E\left(\bar{c}_{ni}\bar{c}_{nj}|s_{ij}|\right)^2  \\
&= \mathcal{O}\left(\frac{1}{n}\right)
\end{align}
\end{mysubequations}  
To go from \eqref{l3var_eq}a to \eqref{l3var_eq}b $\int |s_{ij} s_{jm}|<f_c$ (cauchy-schwartz inequality), P5, P3 are used. To go from  \eqref{l2var_eq}b to  \eqref{l2var_eq}c $\int g(x) < \infty$ due to plancheral, P5 and $\int |s_{ij} g(x_i)|<\sqrt{f_c}$ (cauchy-schwartz inequality)  are used.  
Therefore, again by chebyshev inequality and Borel-Cantelli lemma   $A_{n^2}\overset{a.s.}\to \lim_{n\to\infty}E(A_n)$. Now, consider integer $k$ such that $k^2\leq n\leq (k+1)^2$, as $n^2A_n$  is monotonically increasing (by definition) this implies:
\begin{mysubequations}
\begin{align}
\frac{k^4}{(k+1)^2}A_{k^2}&\leq A_n \leq \frac{(k+1)^4}{k^4}A_{(k+1)^2} \\
\overset{a.s.}{\to} \lim_{n\to\infty}E(A_n) &\leq A_n \leq \lim_{n\to\infty}E(A_n)
\end{align}
\end{mysubequations}
Now by sandwitch theorem $A_n\overset{a.s.}{\to} \lim_{n\to\infty}E(A_n)$, similarly it can be shown that $B_n\overset{a.s.}{\to} \lim_{n\to\infty}E(B_n)$ and hence $S_n\overset{a.s.}{\to} \lim_{n\to\infty}E(S_n)$. Hence proved. 

Now, Theorems \ref{limitsol_thm} and \ref{convergence_thm} are proven.\\

\   \\
\subsubsection{Proof for Consistency of the BLML Estimator}
\begin{theorem}\label{limitsol_thm}
Suppose that the observations, $x_i$ for $i=1,...,n$ are i.i.d. and distributed as $x_i\sim g^2(x)\in \mathbb{U}(\omega_c)$. Then, $\bar{c}_{\infty i} \triangleq \lim_{n\to \infty} \frac{ng(x_i)}{2f_c} \left( \sqrt{1+\frac{4}{n}\frac{f_c}{g^2(x_i)}}-1\right)$ is a solution to $\bm{\rho_n}(\bold{c)=0}$ in the limit as $n\to \infty$.
\end{theorem} 
\noindent
\textit{Proof:} To prove this theorem, we establish that any equation $\rho_{ni}(\bold{\bar{c}_n})$, indexed by $i$ goes to 0 almost surely as $n\to \infty$ as follows:


\begin{mysubequations}\label{rho_eq}
\begin{align}
\rho_{ni}(\bold{\bar{c}_n})&=\frac{1}{n}\sum_{j\neq i}s_{ij}\bar{c}_{nj} + \frac{\bar{c}_{ni} f_c}{n}-\frac{1}{\bar{c}_{ni}} \ \forall i=1,\cdots,n \\
&\overset{a.s.}\to g(x_i)-g(x_i)=0 \ \forall i=1,\cdots,n
\end{align}
\end{mysubequations}
In moving from \eqref{rho_eq}a to \eqref{rho_eq}b (P1) and (P7) are used. \eqref{rho_eq}b, show that each of the $\rho_{ni}(\bar{c}_{n})\ \forall i$ goes to 0 in probability. Therefore, 

\begin{myequation}
\lim_{n\to\infty}\rho_{ni}(\bold{\bar{c}_{n}})=0 \ \forall i=1,\cdots,n 
\end{myequation}

This proves Theorem \ref{limitsol_thm}. Note that one may naively say that $\lim_{n\to\infty}\bar{c}_{ni} = \frac{1}{g(x_i)} \ \forall \ i=1,\cdots,n$ (see (P2)). However, this is not true because even for large $n$ there is a finite probability of getting at least one $g(x_i)$ which is so small such that $\frac{1}{ng^2(x_i)}$ may be finite, and hence $\lim_{n\to\infty}\bar{c}_{ni}$  cannot be calculated the usual way. Therefore, it is wise to write down $\bar{c}_{\infty i}\triangleq \lim_{n\to\infty}\bar{c}_{ni}$ as a solution to \eqref{results_eq}, instead of $\frac{1}{g(x_i)}$. \\ 

\begin{theorem}\label{convergence_thm}
Suppose that the observations, $x_i$ for $i=1,...,n$ are i.i.d. and distributed as $x_i\sim f(x) \in \mathbb{U}(\omega_c)$ and $f(x)>0 \ \forall x$. Let, $f_\infty (x)\triangleq \lim_{n\to\infty} \left(\frac{1}{n}\sum_{i=1}^n\bar{c}_{\infty i}\frac{\sin(\pi f_c(x-x_i))}{\pi(x-x_i)}\right)^2$, then $\int \left(f(x)-f_\infty(x)\right)^2 \mathrm{d}x =0$.
\end{theorem}
\noindent 
\textit{Proof:} Let $\hat{g}_{\infty}(x)\triangleq \lim_{n\to \infty}$ $\frac{1}{n}\sum_{i=1}^n \bar{c}_{\infty i} s(x-x_i)$ here $s(x-x_i)\triangleq \frac{\sin(\pi f_c(x-x_i))}{\pi(x-x_i)}$. Therefore the ISE is:
\begin{small}
\begin{mysubequations}\label{ISE_eq}
\begin{align}
ISE&\triangleq  \int\left( \hat{g}_{\infty}^2(x)-g^2(x)\right)^2 \mathrm{d}x \\
&= \int \left( \hat{g}(x)-\hat{g}_{\infty}(x)\right)^2\left(\hat{g}_{\infty}(x)+g(x)\right)^2 \mathrm{d}x\\
&\leq 4f_c \int (\hat{g}_{\infty}(x)-g(x))^2 \mathrm{d}x\\
&= 4f_c \int \lim_{n\to\infty} \left(\frac{1}{n}\sum \bar{c}_{n j} s(x-x_j)-g(x)\right)^2 \mathrm{d}x\\
&\leq 4f_c \liminf_{n\to\infty} \int  \left(\frac{1}{n}\sum \bar{c}_{n j} s(x-x_j)-g(x)\right)^2 \mathrm{d}x\\
&\overset{a.s.}\to 0
\end{align}
\end{mysubequations}
\end{small}

To go from \eqref{ISE_eq}b to \eqref{ISE_eq}c, the inequality $(g(x)+\hat{g}(x))^2\leq 4f_c$ if $\hat{g},g \in \mathbb{V}$ is used (see Theorem \ref{maxval_thm}). To go from \eqref{ISE_eq}c to \eqref{ISE_eq}d, $\hat{g}_\infty(x)$ is expanded. To go from \eqref{ISE_eq}d to \eqref{ISE_eq}e, Fatou's lemma \cite{Royden2010} is invoked as the function inside the integral is non-negative and measurable. In particular, due to (P6), $\phi_n(x)=\frac{1}{n}\sum  \bar{c}_{n j} s(x-x_j) -g(x) \overset{a.s.}{\rightarrow} E(\bar{c}_{n j} s(x-x_j)) -g(x)=0$, which establishes the point-wise convergence of $\phi_n^2(x)$ to $0$. Hence, $\lim$ can be safely replaced by $\liminf$ and Fatou's lemma can be applied. To go from \eqref{ISE_eq}e to \eqref{ISE_eq}f, \eqref{l2norm_eq} is used.


Hence proved.\\

\begin{theorem}\label{globalmax_thm}
Suppose that the observations, $x_i$ for $i=1,...,n$ are i.i.d. and distributed as $x_i\sim f(x) \in \mathbb{U}(\omega_c)$. Then, the KL-divergence between $f(x)$ and $f_\infty(x)$ is zero and hence $\bold{\bar{c}_{\infty}}$ is the solution of \eqref{BLML_eq2} in the limit $n\to \infty$. Therefore, the BLML estimator $\hat{f}(x)=f_\infty(x)=f(x)$ in probability.
\end{theorem}
\noindent 
\textit{Proof:} Consider $\{x_1,\cdots,x_n\}$ to be a member of typical set \cite{Cover1991}. Then the KL-divergence between $f(x)$ and $f_\infty (x)$ can be bounded as:

\begin{mysubequations}\label{KL_eq}
\begin{align}
0&\leq E\left(\log\left(\frac{f(x)}{f_\infty(x)}\right)\right)=\lim_{n\to\infty}\frac{1}{n}\sum_{i=1}^n\log\left(\frac{g^2(x_i)}{g^2_\infty(x_i)}\right) \\
& \leq \lim_{n\to\infty} \frac{2}{n} \sum_{i=1}^n \log\left(  |g(x_i)\bar{c}_{\infty i}|\right) \\
&  \leq 0 
\end{align}
\end{mysubequations} 
To go from  \eqref{KL_eq} a to \eqref{KL_eq} b, definition of $g_\infty$ and P7 is used. To go from  \eqref{KL_eq} b to \eqref{KL_eq}c, (P5) is used. 

Therefore, the KL divergence between $\hat{f}_\infty(x)$ and the true pdf is $0$ and hence $\hat{f}_\infty(x)$ should also maximizes the likelihood function.  Finally, $\bold{\hat{c}=\bar{c}_\infty}$ or $\bold{\hat{c}=-\bar{c}_\infty}$. The negative solution can be safely ignored by limiting only to positive solutions. Hence Proved.\\

\vspace{1 in}

\begin{theorem}\label{Trivial_thm}
If $g^2(x)=f(x) \in \mathbb{U}(\omega_c)$ such that $f(x)>0 \ \  \forall \ x \in \mathbb{R},$ then $g(x)>0 \ \  \forall \ x \in \mathbb{R},$ and  the asymptotic solution of \eqref{BLML_eq2} lies in the orthant with indicator vector $c_{0i}=1\ \forall i=1,\cdots,n$.  
\end{theorem}
\noindent
\textit{Proof:} $g \in \mathbb{V}(\omega_{c})$ as $g^2 \in \mathbb{U}(\omega_c)$. Therefore $g(x)$ is bandlimited and hence continuous. Now, assume that $\exists\  x_1,x_2\in\mathbb{R}$ such that $g(x_1)>0 \ \text{and}\  g(x_2)<0$. Due to continuity of $g$ this would imply that $\ \exists\  x_3,\  x_1<x_3<x_2$ such that $g(x_3)=f(x_3)=0$. This is a contradiction as $f(x)>0 \ \forall x\in\mathbb{R}$. Therefore, either $g(x)<0\ \forall \ x\in\mathbb{R}$ or equivalently $g(x)>0\  \forall\  x\in\mathbb{R}$.  Now, by Theorems  \ref{limitsol_thm} and \ref{globalmax_thm}, $c_{0i}=sign(\hat{c}_i)=sign(c_{\infty i})=sign(g(x_i))=1 \ \forall i=1\cdots n$ asymptotically. Hence proved.

\subsection{Generalization of BLML estimator to Joint Pdfs}
BLML estimator for joints pdfs can be found in a very similar way as it is found for one dimensional pdfs. The only change occurs while defining \eqref{bi_eq}, where one needs to define multidimensional $b_i's$ such that
\begin{equation} \label{bigen_eq}
b_i(\bm{\omega}) \triangleq  \left\lbrace \begin{array}{cc} e^{-j\bm{\omega}^T \bold{x}_i} & \forall  \  |\bm{\omega}| \leq \frac{\bm{\omega}_c}{2} \\
0 &  o.w. \end{array} \right\rbrace,
\end{equation}
\noindent 
inverse Fourier transform of which gives a multidimensional $\sinc_\bold{f_c}(\cdot)$ function.

\subsection{{\bf {\small \texttt{BLMLQuick}}} Algorithm}
Consider a function $\bar{f}(x)$ such that:
\begin{myequation}
\bar{f}(x)=f_s \int_{x-\frac{0.5}{f_s}}^{x+\frac{0.5}{f_s}} f(\tau) d\tau
\end{myequation}

\noindent
where $f\in\mathbb{U}(\omega_c)$ and $f_s>2f_c$ is the sampling frequency. It is easy to verify that $\bar{f}(x)$ is also a pdf and $\bar{f}\in\mathbb{U}(\omega_c)$. Now consider samples $\bar{f}[p]=\bar{f}(p/f_s)$, clearly these samples are related to $f(x)$ as: 

\begin{myequation}
\bar{f}[p]=\int_{\frac{p-0.5}{f_s}}^{\frac{p+0.5}{f_s}} f(x) dx
\end{myequation}

Further consider $\bar{x}_i$'s computed by binning from $x_i$'s, $n$ i.i.d observations of r.v. $x\sim f(x)$, as:
\begin{myequation}
\bar{x}_i=f_s\lfloor \frac{x_i}{f_s} -0.5 \rfloor
\end{myequation}

where $\lfloor\rfloor$ is the greatest integer function. Note that $\bar{x}_i$ are the i.i.d. observations from $\tilde{f}(x)=\sum_p \bar{f}[p]\delta\left(x-\frac{p}{f_s}\right)$. Now since $f_s>2f_c,$ the BLML estimate for $\tilde{f}(x)$ should converge to $\bar{f}(x)$ due to Nyquist's Sampling Theorem \cite{Marks1991}. This estimator is called \texttt{BLMLQuick}. Assuming that the rate of convergence for BLML is $\mathcal{O}(n^{-1}),$ then if $f_s$ is chosen such that $||f-\bar{f}||_2^2=\mathcal{O}(n^{-1}),$ the \texttt{BLMLQuick} will also converge with $\mathcal{O}(n^{-1})$. This happens at $f_s=f_cn^{0.25}> f_c$ also $f_s>2f_c$ if $n>16$.

\subsubsection*{Implementation and Computational Complexity}

Before implementing \texttt{BLMLQuick} and computing its computational complexity, the following theorem is first stated and proved.
\vspace{0.1in}

\noindent
\begin{theorem}\label{maxobs_thm}
Consider $n$ i.i.d observations $\{x_i\}_{i=1}^n$ of random variable $x$ with pdf having $\frac{1}{|x|^r}$ tail. Then 
\begin{myequation}
\Pr\left(\min(\{x_i\}_{i=1}^n)<-\left(\frac{n}{(r-1)\epsilon}\right)^{\frac{1}{r-1}}\right)\simeq 1-e^{-\epsilon}\simeq \epsilon
\end{myequation}

for large $n$.

\end{theorem}

\vspace{0.1in}

\textit{Proof}: For $n$ i.i.d observations $\{x_i\}_{i=1}^n$ of random variable $x$ with cumulative distribution function $F(x)$, it is well known that :
\begin{mysubequations}
\begin{align}
\Pr(\min(\{x_i\}_{i=1}^n)<x)&=1-(1-F(x))^n\\
&\simeq 1-e^{-nF(x)}\  \forall F(x)<0.5\\
&\simeq 1-e^{-\frac{n}{(r-1)|x|^{r-1}}} \  \forall F(x)<0.5
\end{align}
\end{mysubequations}

\noindent
substituting $x= -\left(\frac{n}{(r-1)\epsilon}\right)^{\frac{1}{r-1}}$ above proves the result.

Finally, due to duplicity in ${\bar{x}_i}$ $i=1,\dots,n$, they can be written concisely as $[\bar{x}_b,n_b]$, $b=1,\dots,B$ where ${\bar{x}_b}$ are unique values in ${\bar{x}_i}$ and $n_b$ is duplicity count of $\bar{x}_b$. Now it can be observed that $B\leq (\max(x_i)-\min(x_i))f_s \leq \mathcal{O}_p(n^{\frac{1}{r-1}})f_s$, if the true pdf has tail that decreases as $\frac{1}{|x|^r}$ (Theorem \ref{maxobs_thm}). 

Now we are set to implement BLML quick using following steps:

\begin{itemize}
\item Compute $\{\bar{x}_b,n_b\}_{b=1}^B$ from $\{x_i\}_{i=1}^n$. Computational complexity of $\mathcal{O}(n)$.
\item Sort $\{\bar{x}_b,n_b\}_{b=1}^B$ and construct $\bold{S}:s_{ab}=s(\bar{x}_a-\bar{x}_b) \ \forall\ a,b=1,\dots,B$ and $\bold{\bar{S}}=\bold{S}*diag(\{n_b\}_{b=1}^B)$. Note that $\bold{S}$ is block-Toeplitz matrix (Toeplitz arrangements of blocks and each block is Toeplitz) \cite{Akaike1973}.  Computational complexity of $\mathcal{O}(B^2)$.
\item Use convex optimization algorithms to solve $\rho_n(\bold{c})=0$. Newton's method should take a finite number of iterations to reach a given tolerance $\epsilon$ since the cost function is self concordant \cite{Boyd2004}. Therefore, the computational complexity of optimization is same as the computational complexity of one iteration. The complexity of one iteration is the same as the complexity of calculating  
\begin{mysubequations}
\begin{align}
&\left(diag\left(\{1/c_b^2\}_{b=1}^B\right)+\bold{S} \times diag\left(\{n_b\}_{b=1}^B\right)\right)^{-1} \\ &~~~~~~~~~~~~~=\left(diag\left(\{1/(c_b^2n_b)\}_{b=1}^B\right)+\bold{S}\right)^{-1}diag\left(\{n_b\}_{b=1}^B\right)^{-1}
\end{align}
\end{mysubequations}

As $diag\left(\{1/(c_b^2n_b)\}_{b=1}^B\right)+\bold{S}$ is also block-Toeplitz structure, the Akaike algorithm \cite{Akaike1973} can be used to evaluate each iteration of Newton's method  in $\mathcal{O}(B^2)$.\\
 Note: Simulations show that $\bold{S}$ can be approximated accurately (to machine accuracy) by a low rank matrix e.g., $R=20$ for $B=1000$, therefore the inversion can be performed in $\mathcal{O}(R^2+RB)$. Further, in some cases one may end up with a large $B$ (e.g. if true pdf has heavy tails) so that storing the Hessian matrix becomes expensive. In such cases, a quasi Newton or gradient descent can be used which compute BLML estimator fairly quickly.  
\item Evaluate \texttt{BLMLQuick} estimate $f(x)=(\frac{1}{n}\sum_{b=1}^{B}n_bc_bs(x-x_b))^2$ at $l$ given points. Computational complexity of $\mathcal{O}(Bl)$.

The total computational complexity is $\mathcal{O}(n+B^2+lB)$. Substituting $B\leq\mathcal{O}\left(n^{\frac{1}{r-1}}\right)f_s\leq\mathcal{O}\left(f_cn^{\frac{1}{r-1}+0.25}\right)$, gives the total computational complexity \\ $\mathcal{O}\left(n+f_c^2n^{\frac{2}{r-1}+0.5}+f_cln^{\frac{1}{r-1}+0.25}\right)$. 

\end{itemize}


\end{article}
\end{document}